\documentclass[10pt,twocolumn,letterpaper]{article}

\usepackage[accsupp]{axessibility}  % Improves PDF readability for those with disabilities.

\usepackage{wacv}              % To produce the CAMERA-READY version
\usepackage{graphicx}
\usepackage{amsmath}
\usepackage{amssymb}
\usepackage{booktabs}

\usepackage{subcaption}
\usepackage{multirow}
\usepackage{pifont}
\usepackage{lineno}
\usepackage{verbatim}

\usepackage[pagebackref,breaklinks,colorlinks]{hyperref}

\usepackage[capitalize]{cleveref}
\crefname{section}{Sec.}{Secs.}
\Crefname{section}{Section}{Sections}
\Crefname{table}{Table}{Tables}
\crefname{table}{Tab.}{Tabs.}

\def\wacvPaperID{1233} % *** Enter the WACV Paper ID here
\def\confName{WACV}
\def\confYear{2024}

\begin{document}

\title{Lightweight Portrait Matting via Regional Attention and Refinement}

\author{Yatao Zhong\\
Microsoft\\
{\tt\small yazhong@microsoft.com}
\and
Ilya Zharkov\\
Microsoft\\
{\tt\small zharkov@microsoft.com}
}
\maketitle

%%%%%%%%% ABSTRACT
\begin{abstract}
We present a lightweight model for high resolution portrait matting. The model does not use any auxiliary inputs such as trimaps or background captures and achieves real time performance for HD videos and near real time for 4K. Our model is built upon a two-stage framework with a low resolution network for coarse alpha estimation followed by a refinement network for local region improvement. However, a naive implementation of the two-stage model suffers from poor matting quality if not utilizing any auxiliary inputs. We address the performance gap by leveraging the vision transformer (ViT) as the backbone of the low resolution network, motivated by the observation that the tokenization step of ViT can reduce spatial resolution while retain as much pixel information as possible. To inform local regions of the context, we propose a novel cross region attention (CRA) module in the refinement network to propagate the contextual information across the neighboring regions. We demonstrate that our method achieves superior results and outperforms other baselines on three benchmark datasets while only uses $1/20$ of the FLOPS compared to the existing state-of-the-art model.
\end{abstract}

%%%%%%%%% BODY TEXT
\section{Introduction}
Image matting is one of the most studied topics in computer vision. Formally a matting problem is formulated as
\begin{equation}
\label{eq:matting}
I=\alpha F + (1 - \alpha) B.
\end{equation}
The goal is to solve for the alpha matte $\alpha$, but the foreground $F$ and background $B$ are also unknown. Therefore, this is a highly under-constrained problem, which oftentimes requires some priors. One commonly used prior is a user provided trimap \cite{levin2007closed, aksoy2017designing, chen2013knn, chuang2001bayesian, sun2004poisson, xu2017deep, lutz2018alphagan, li2020natural, lu2019indices, forte2020f}, where each pixel is categorized as ``definite foreground'', ``definite background'' or ``unknown''. However, trimaps require user interaction and are time-consuming to obtain, hence difficult to be deployed in a fully automated system. Another recently proposed prior is an additional background image \cite{sengupta2020background}. However, capturing a second image under the same conditions (e.g., lighting and shadow) is not always possible and the background image is only useful if it is well aligned with the input image. There have also been efforts to remove all auxiliary inputs and predict the alpha mattes directly from input images \cite{chen2018semantic, zhang2019late, qiao2020attention, li2021privacy}. Approaches of this type are typically learning based and have been demonstrated to perform reasonably well even without any priors provided.

\begin{figure}[t]
    \centering
    \includegraphics[width=\linewidth]{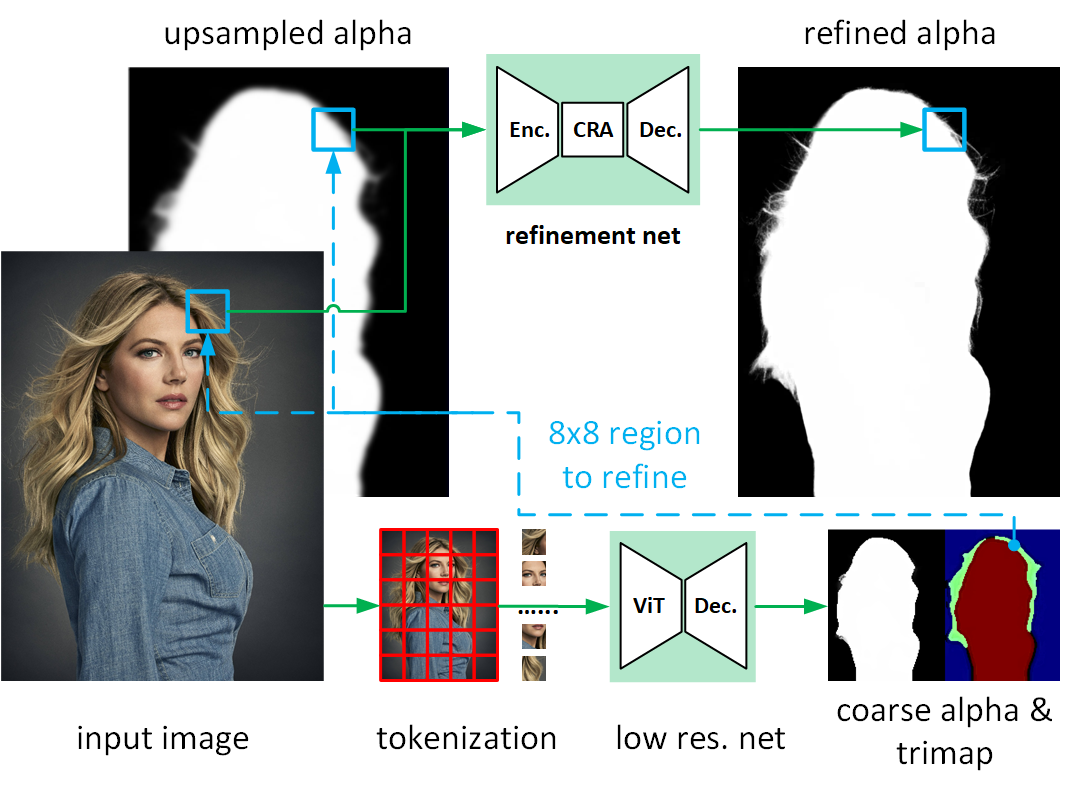}
    \caption{An overview of our method. 1) An input image is first tokenized before being fed to a low resolution network that consists of a ViT backbone and a decoder. 2) Coarse alpha is upsampled to full resolution and concatenated with the input image. 3) Regions of uncertainty are selected from the estimated trimap and cropped from the concatenated RGBA image. 4) Cropped regions run through a refinement network that features a cross region attention (CRA) module to obtained the refined alpha.}
    \label{fig:pull_fig}
\end{figure}

Nevertheless, all of these methods operate at full resolution, making them extremely compute-intensive and impossible to be deployed in real applications for high resolution portrait matting (e.g., HD and 4K). In this work, we aim at reducing the computation while also retaining the matting quality. We present a lightweight model that estimates the alpha matte directly from the image without any user interactions or auxiliary inputs such as trimaps or background captures. Our model is built upon a prior observation that a portrait alpha matte is dominated by ``definite foreground'' ($\alpha=1$) and ``definite background'' ($\alpha=0$), which can be obtained by upsampling the estimated alpha matte from a low resolution model. Only a few uncertain regions around the boundaries ($0<\alpha<1$) need to be refined. Therefore, the proposed model consists of two stages: an initial stage for low resolution alpha estimation and a second stage for full resolution refinement. Fig. \ref{fig:pull_fig} gives an overview of our method.

However, we find naively adopting the two-stage framework leads to inferior results due to the missing auxiliary inputs which we intentionally eliminate. To address the performance gap, we leverage the vision transformer (ViT) as the backbone in the low resolution model. As opposed to image downsampling, image tokenization in the ViT is a better choice for reducing spatial resolution because it does not lose pixel information. Since the the refinement network operates on extracted local regions, to inform it of the context, a straightforward design choice would be to reuse the upsampled features from the low resolution network \cite{lin2021real}. However, this adds to the compute budget by doing upsampling at high resolution. Therefore, we opt for an inverted process by first extracting local regions followed by gathering the context. To recover the contextual information, we propose a novel cross region attention (CRA) module, which propagates the information across the $k$ nearest neighbors of each region through multi-head attention with a learnable lookup table for relative positional encoding. We demonstrate that, with all the aforementioned designs, our model outperforms other baselines by a large margin on the P3M and PPM datasets. We also show that our model is able to retain the matting quality using only $1/20$ of the FLOPS compared to the existing state-of-the-art model \cite{li2021privacy}.

In summary, our work has the following contributions.
\begin{itemize}
\item We leverage the tokenization step of ViT to reduce spatial resolution while retain the full pixel information for coarse alpha estimation.
%\item We opt for local region extraction followed by contextual information collection to avoid feature upsampling at high resolution to save computation.
\item We invert the order of computing contextual features and extracting local regions to avoid feature upsampling at high resolution to save computation. 
\item We propose a novel cross region attention (CRA) module to capture the contextual information across $k$ nearest neighbors of each region.
\item We conduct extensive experiments to demonstrate the effectiveness and efficiency of our model: achieving the state-of-the-art performance while using minimal FLOPS.
\end{itemize}

\section{Related Work}
\textbf{Traditional matting.} Traditional matting algorithms \cite{levin2007closed, he2010fast, aksoy2017designing, chen2013knn, chuang2001bayesian, sun2004poisson} are derived from the matting equation Eq. \ref{eq:matting}. The goal is to solve for the alpha matte $\alpha$, but at the same time one needs to also solve for the foreground $F$ and background $B$. Since this is an ill-posed problem with only the observed image $I$ being provided, a common practice is to use trimaps as constraints. \cite{chuang2001bayesian} formulates the problem in a Bayesian framework and solves it using maximum a posteriori (MAP) estimation. \cite{sun2004poisson} formulates matting as a problem of solving Poisson equations using matte gradient field. Both end up being an iterative solution. \cite{levin2007closed} proposes the first closed form solution, but their method is memory and compute intensive because the involvement of a large sparse linear system.  \cite{he2010fast} accelerates \cite{levin2007closed} by using large kernel Laplacian and adaptive kernel sizes obtained from KD-tree segmentation on trimaps. Other works \cite{chen2013knn, aksoy2017designing} improve \cite{levin2007closed} by removing the local color line model assumption. They use the global pixel affinities to propagate alpha values in trimaps from known regions to unknown regions.

\textbf{Learning based matting.} With the advance in deep learning, many recent approaches \cite{xu2017deep, lutz2018alphagan, li2020natural, lu2019indices} have shifted to a learning based paradigm, where a model takes the image and trimap as input and learns to predict the alpha matte. DIM \cite{xu2017deep} is the pioneer that leverages deep neural networks in the task of image matting. AlphaGAN \cite{lutz2018alphagan} improves \cite{xu2017deep} by training the model with an adversarial loss. GCA \cite{li2020natural} introduces a guided contextual attention module by computing the correlation between unknown regions. IndexNet \cite{lu2019indices} utilizes learned indices in the decoder for upsampling to guide the matte generation.

\textbf{Trimap-free matting.}
There have also been efforts trying to eliminate the dependence on trimaps. \cite{sengupta2020background, lin2021real} propose to capture an additional background image as an auxiliary input for image matting. SHM \cite{chen2018semantic} predicts the alpha matte by fusing a self-learned trimap and a raw alpha matte. LF \cite{zhang2019late} employs a similar fusion concept by estimating the foreground and background probability maps and blending them with self-learned weights. HATT \cite{qiao2020attention} uses a spatial and channel-wise attention module to integrate low level and high level features. P3M-Net \cite{li2021privacy} adopts a multi-task framework by predicting trimaps and alpha mattes at multi-resolutions and uses a stack of integration modules to exchange feature information.

\section{Method}
The proposed two-stage framework (shown in Fig. \ref{fig:pull_fig}) proceeds as follows. The low resolution network predicts a coarse alpha matte and a coarse trimap. Next we extract uncertain regions using the coarse trimap and crop the selected regions from the input image and upsampled coarse alpha. The cropped patches then pass through the refinement network to obtain the refined alpha patches. Finally, we replace the refined alpha patches back in the upsampled coarse alpha to complete the full alpha. Below we illustrate each of the steps and explain our design choices. For brevity of writing, we use the notation $\mathfrak{R}_x$ to refer to a resolution at $\frac{1}{x}$ of the full resolution.

\subsection{ViT as Low Resolution Backbone} \label{sec:vit_for_low_res}
Different from prior works that use additional inputs such as background captures \cite{lin2021real,sengupta2020background} or trimaps \cite{xu2017deep,lutz2018alphagan,li2020natural,lu2019indices,forte2020f}, our models tackles portrait matting without any auxiliary inputs, which is significantly more challenging. In fact, as we will show in the experiments, simply adopting a CNN architecture results in poor matting quality if not given auxiliaries. We argue that, even for coarse alpha estimation, a higher input resolution could be beneficial to the overall better quality. Nonetheless, higher resolution inevitably adds more computation. This motivates us to think how we can improve it without resorting to increased compute budget. We therefore propose to transform an image $I \in \mathbb{R}^{H \times W \times C}$ to a grid of non-overlapping patches of $\mathbb{R}^{\frac{H}{P} \times \frac{W}{P} \times P^2 C}$, where $\frac{H}{P} \times \frac{W}{P}$ corresponds to the grid resolution and $P^2 C$ is the channel dimension. This is often referred to as \textit{pixel-unshuffle} or \textit{space-to-depth}. As opposed to downsampling, pixel-unshuffle preserves the original pixel information when reducing the spatial resolution.

\setlength{\tabcolsep}{0.006\textwidth} % set cell horizontal padding
\begin{table}[h]
    \small
    \centering
    \caption{Results of different downsampling and pixel-unshuffle strategies. $d$ is the downsampling rate and $p$ is the pixel-unshuffle patch size. We evaluate the models on the P3M-500 test data. For more details about the test data, please refer to Sec. \ref{sec:exp_setup}.}
    \label{tab:trial_exp}
    \begin{tabular}{c|c|c|c|c|c|c}
    \toprule
    \multicolumn{3}{c|}{}  & \multicolumn{2}{c|}{P3M-500-NP} & \multicolumn{2}{c}{P3M-500-P} \\
    \hline
    No. & Method & FLOPS & SAD & Grad & SAD & Grad \\
    \hline
    % e40
    A & Resnet-50, $d$=4 & 24.7G & 14.25 & 12.60 & 14.11 & 14.69 \\
    \hline
    % e41
    B & Resnet-50, $d$=2 & 59.9G & 13.43 & 11.77 & 12.05 & 12.58 \\
    \hline
    % e39
    C & Resnet-50, $p$=4 & 28.2G & 15.21 & 12.71 & 16.10 & 15.22 \\
    \hline
    % h2_1
    D & Swin-T, $d$=2, $p$=8 & 18.6G & 10.89 & 10.72 & 10.39 & 12.73 \\
    \bottomrule
    \end{tabular}
\end{table}

To verify the advantage of pixel-unshuffle over downsampling, we test several models with different pixel-unshuffle and downsampling strategies and evaluate them on a benchmark dataset. We summarize their SAD (sum of absolute difference) and Grad (gradient difference) in Tab. \ref{tab:trial_exp}.  Resnet-50 \cite{he2016deep, he2016identity} is used as the low resolution backbone for models A, B and C. A sequence of upsampling and conv layers are used in the decoder to keep the low resolution output at $\mathfrak{R}_8$. All models share the same refinement stage, which we will discuss later in Sec. \ref{sec:refine_net} and \ref{sec:attn}.  

Comparing A and B, we can see that increasing the resolution of the low resolution network improves the accuracy. Nevertheless, this requires considerably more compute budget. To retain accuracy without adding more computation, we resort to pixel-unshuffle (C). However, we see C underperforms A. This seemingly contradicts the hypothetical strength of using pixel-unshuffle, but we argue that the performance drop, in fact, can be explained by the usage of large kernels. Many prior arts \cite{he2016deep, he2016identity, simonyan2014very, szegedy2015going, zhang2018residual} have demonstrated the success of using small kernels because they help preserve the locality and translation invariance of CNNs. Large kernels break these nice properties, making CNNs suffer from poor generalization. In the case of C in Tab. \ref{tab:trial_exp}, Resnet-50 already starts with a relatively large kernel (7$\times$7). When used with pixel-unshuffle with a patch size of 4, the effective kernel size of the first layer becomes 28$\times$28, which is obviously too large to be applied to a CNN.

Due to limitation of CNNs with large kernels, we opt for ViT as the low resolution backbone. ViT comes naturally a better choice for low resolution prediction because the first step of ViT --- image tokenization --- is equivalent to pixel-unshuffle, which can effectively reduce the spatial resolution while retain the full pixel information. Specifically, we choose the Swin-T \cite{liu2021swin} as the low resolution backbone. As shown in Tab. \ref{tab:trial_exp}, model D uses a downsampling rate of 2 followed by pixel-unshuffle with a patch size of 8, which results in a $\times$16 reduction in resolution. With the proposed design principal, model D achieves a remarkable improvement in terms of both accuracy and FLOPS. 

\subsection{Refinement Stage} \label{sec:refine_net}
Modern neural network architectures for dense prediction tasks typically rely on a pyramid of upsampled features for global information. Similarly, for a refinement network to receive contextual information, the most straightforward idea would be to upsample the features to input resolution before extracting the refinement regions. This way the cropped regions are informed of their context. However, upsampling at high resolution (e.g. HD or 4K) is both memory and compute intensive. We propose an alternative to eliminate the heavy feature upsampling.

Our refinement stage avoids reusing any deep features from the low resolution network. The low resolution network only predicts a coarse alpha matte and a coarse trimap. Like a traditional trimap, the predicted trimap has three classes: ``definite foreground'', ``definite background'' and ``uncertain''. We encode it as a 3-channel softmax output. Since no ground truth trimaps are available at training time, we apply morphological operations with heuristics to create the target trimaps from the ground truth alpha mattes. At inference time, we select the pixels predicted as ``uncertain'' as the regions of interest to be refined. 

In the refinement stage, we first upsample the coarse alpha matte to full resolution. This op is lightweight compared to the heavy feature upsampling. The upsampled alpha matte is concatenated with the input image to form a 4-channel RGBA image. With the selected ``uncertain'' pixels from the trimap, we locate the corresponding 8$\times$8 regions in the RGBA image, which are cropped and fed to a tiny refinement network consisting of an encoder and a decoder. At last, we replace the respective 8$\times$8 regions in the upsampled alpha with the refined crops to obtain the final alpha. Fig. \ref{fig:refine_net} visualizes how the cropped regions run through the entire refinement stage.

\subsection{Cross Region Attention} \label{sec:attn}
 The aforementioned refinement stage only receives 8$\times$8 local regions as input, inevitably losing the context. Thus a mechanism is needed to recover the context after region extraction. We therefore propose a novel cross region attention (CRA) module to capture the contextual information across the neighboring regions. CRA is inspired by the multi-head attention \cite{vaswani2017attention, dosovitskiy2020image, liu2021swin}, but instead of consuming a sequence or regular grid of tokens, it operates on the $k$ nearest neighbors (KNN) of a central token. Our proposed mechanism inverts the order of context collection and region extraction and has the advantage of eliminating the heavy feature upsampling as discussed in Sec. \ref{sec:refine_net}. Below we use ``region'' to refer to ``token'' since an extracted region is effectively a token.

\begin{figure*}[t]
  \centering
  \hspace{0.04\textwidth}
  \begin{subfigure}[b]{0.7\textwidth}
      {
        \includegraphics[width=\textwidth]{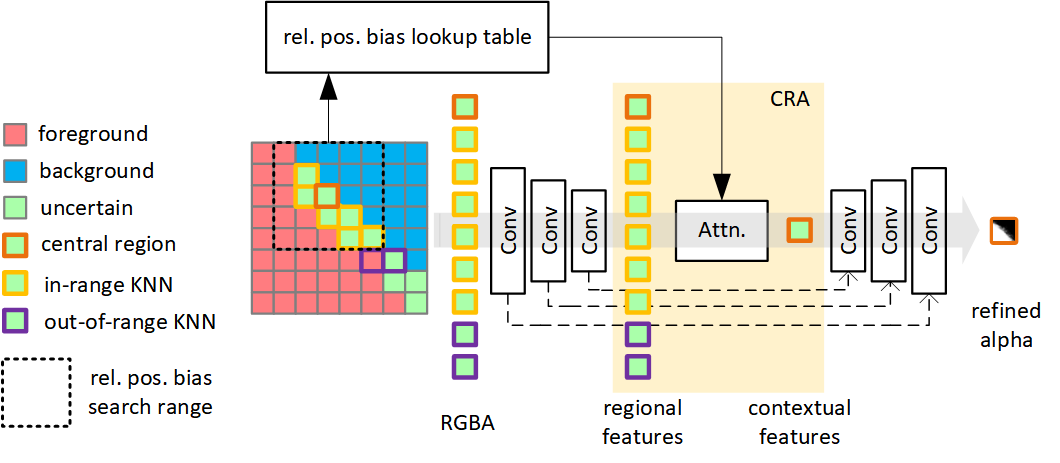}
        %\vspace{-0.04\hsize}
        \caption{}
        \label{fig:refine_net}
      }
  \end{subfigure}
  \hfill
  \begin{subfigure}[b]{0.21\textwidth}
      {
        \includegraphics[width=\textwidth]{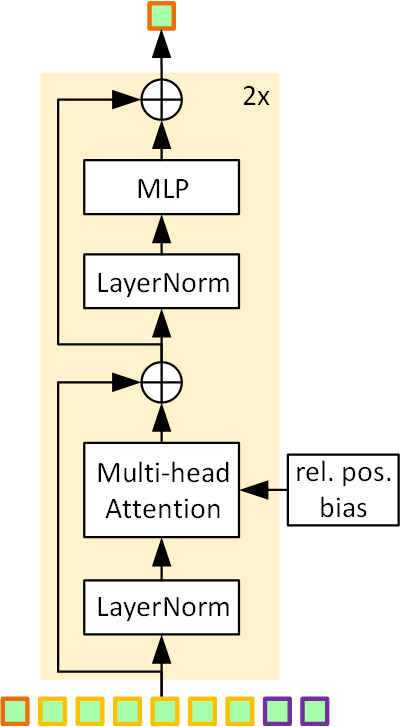}
        %\vspace{-0.04\hsize}
        \caption{}
        \label{fig:CRA}
      }
  \end{subfigure}
  \hspace{0.02\textwidth}
  \caption{A visualization of the refinement stage. Each $\square$ corresponds to a region of 8$\times$8 pixels at full resolution. The example uses 8 nearest neighbors and a search range $s=3$ for relative positional bias. (a) visualizes the KNNs of a central region and the identification of in-range and out-of-range neighbors. A central region obtains its contextual features by aggregating the features from its KNNs through the cross region attention (CRA) module. (b) shows the structure of the attention block in CRA.}
  \label{fig:refine_net_and_CRA}
\end{figure*}

\textbf{KNN extraction}. After identifying all ``uncertain'' regions from the trimap, for each region, we find the closest $k$ regions as its KNNs (under the metric of Euclidean distance). We do pairwise comparisons at training time, but employ KD-tree as a faster search algorithm at inference time. The left part of Fig. \ref{fig:refine_net} visualizes the locations of a region's KNNs.

\begin{figure}[t]
    \centering
    \includegraphics[width=0.55\linewidth]{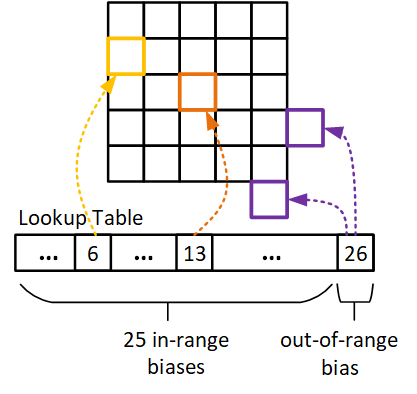}
    %\vspace{-0.02\hsize}
    \caption{An illustration of relative positional bias encoding. This is an example with search range $s$=3, which ends up with a 5$\times$5 search window and a lookup table of 25 biases for in-range positions and one extra bias for any out-of-range positions. The orange square denotes the center, which is encoded with the bias at the 13th slot. The yellow square denotes an in-range sample encoded with the 6th slot. The purple squares are two out-of-range samples, whose positional biases are given by the last table entry.}
    \label{fig:pos_bias}
\end{figure}

\textbf{Relative positional bias}. The KNNs can potentially be scattered anywhere around the central region and distributed on a non-regular grid, so we need a way to encode their relative positions. We define a search range $s$ on the image space. The relative positions on each axis are supposed to be in $[-s+1, s-1]$. This ends up with $(2s-1)^2$ possible relative positions within the search range.       
We encode the relative positions with a learnable lookup table $P \in \mathbb{R}^{(2s-1)^2+1}$. The first $(2s-1)^2$ entries encode all possible in-range positions. The last entry encodes any out-of-range positions. Each entry is used as the relative positional bias $B$ in the attention formula:
\begin{equation}
    \text{Attention}(Q, K, V) = \text{Softmax}(\frac{QK^T}{\sqrt{d}}+B)V \text{,}
\end{equation}
where $d$ is the feature dimension; $Q$, $K$ and $V$ are the query, key and value respectively. Fig. \ref{fig:pos_bias} illustrates how the relative positional biases are encoded with a lookup table.

\textbf{Cross region attention}. After we obtain the features of all extracted regions from the refinement network's encoder, we locate the KNNs of each region and query their relative positional biases from the lookup table $P$. Let $f_i \in \mathbb{R}^d$ denote the feature of region $i$ and $\{i_1, i_2, \cdots, i_k\}$ denote the $k$ nearest neighbors of region $i$. For each region $i$, we feed the features $[f_i, f_{i_1}, f_{i_2}, \cdots, f_{i_k}] \in \mathbb{R}^{(k+1) \times d}$ of this region and its KNNs, along with their relative positional biases $B \in \mathbb{R}^{k+1}$, to two consecutive attention blocks, shown in Fig. \ref{fig:CRA}. Note that the second block does not need to do pairwise attention $QK^T$ across all $k+1$ regions. Instead, it only computes the attention between the central region and its KNNs by $Q_0K^T$, where $Q_0$ is the query of the central region $i$. Finally, the output feature goes to the refinement network's decoder to obtain the refined alpha matte of region $i$.

\subsection{Training}
We train the low resolution network and refinement network end-to-end at the same time. During training, we apply various data augmentation strategies such as horizontal flipping, cropping and affine transformation as well as color adjustment in hue, saturation and brightness. Since the Swin-T backbone inherently uses an effective output stride of 32 and a window size of 7 for its window attention, with the original implementation \cite{liu2021swin} the input image size is expected to be multiples of $7\times32=224$. We make a modification to accommodate input images of arbitrary sizes (but no less than 224$\times$224) by padding any intermediate feature maps with zeros if their spatial sizes are not already dividable by 7 and masking out the padded regions when computing the attention. At training time, all images are resized to 896$\times$896, but at inference time, the model accepts images of arbitrary resolutions. We refer readers to the supplementary material for a full list of training losses.

\setlength{\tabcolsep}{0.015\textwidth} % set cell horizontal
\begin{table*}[th]
    \small
    \centering
    \caption{Quantitative results on the P3M-500 tet data. $\dagger$ indicates that a trimap is used.}
    \label{tab:main_exp}
    \begin{tabular}{c|c|c|c|c|c|c|c|c|c|c}
    \toprule
    \multicolumn{3}{c|}{}  & \multicolumn{4}{c|}{P3M-500-NP} & \multicolumn{4}{c}{P3M-500-P} \\
    \hline
    \multicolumn{2}{c|}{Method} & GFLOPS & SAD & SAD-T & Grad & Conn & SAD & SAD-T & Grad & Conn \\
    \hline
    %\multicolumn{2}{c|}{LF$^\dagger$ \cite{zhang2019late}} & 7190.0 & 32.59 & 14.53 & 31.93 & 19.50 & 42.95 & 12.43 & 42.19 & 18.80  \\
    %\hline
    %\multicolumn{2}{c|}{HATT$^\dagger$ \cite{qiao2020attention}} & 4264.3 & 30.53 & 13.48 & 19.88 & 27.42 & 25.99 & 11.03 & 14.91 & 25.29 \\
    %\hline
    %\multicolumn{2}{c|}{SHM$^\dagger$ \cite{chen2018semantic}} & 1943.3 & 20.77 & 9.14 & 20.30 & 17.09 & 21.56 & 9.14 & 21.24 & 17.53 \\
    %\hline
    %\multicolumn{2}{c|}{AIM$^\dagger$ \cite{li2021deep}} & 487.4 & 15.50 & 10.16 & 14.82 & 18.03 & 13.20 & 8.84 & 12.58 & 17.75 \\
    %\hline
    \multicolumn{2}{c|}{DIM$^\dagger$ \cite{xu2017deep}} & 791.6 & 5.32 & 5.32 & 4.70 & 7.70 & 4.89 & 4.89 & 4.48 & 9.68 \\
    \hline
    \multicolumn{2}{c|}{P3M-Net \cite{li2021privacy}} & 364.9 & 11.23 & 7.65 & 10.35 & 12.51 & 8.73 & 6.89 & 8.22 & 13.88 \\
    \hline
    \multirow{2}{*}{MODNet \cite{ke2022modnet}} & 512x512 input & 15.7 & 20.20 & 12.48 & 16.83 & 18.41 & 30.08 & 12.22 & 19.73 & 28.61 \\
    \cline{2-11}
    & fullres input & 103.2 & 63.74 & 13.56 & 25.75 & 62.69 & 95.47 & 13.70 & 37.28 & 94.86 \\
    \hline
    % e22
    %\multirow{3}{*}{BGMv2 \cite{lin2021real}} & MobileNet-V2 & 7.7 & 21.27 & 8.16 & 14.16 & 18.91 & 21.28 & 7.58 & 17.06 & 19.48 \\
    %\cline{2-11}
    % e23/300k iters
    \multirow{2}{*}{BGMv2 \cite{lin2021real}} & Resnet-50 & 26.5 & 16.72 & 7.55 & 13.00 & 15.39 & 15.70 & 7.23 & 15.54 & 14.71 \\
    \cline{2-11}
    % e24_1/600k iters
    & Resnet-101 & 33.9 & 15.66 & 7.72 & 12.42 & 14.65 & 13.90 & 7.23 & 14.69 & 13.13 \\
    \hline
    % h7
    \multicolumn{2}{c|}{Ours} & 19.0 & 10.60 & 6.83 & 10.78 & 9.77 & 10.04 & 6.44 & 12.65 & 9.41 \\
    \bottomrule
    \end{tabular}
\end{table*}

\section{Experiments} 
\subsection{Experiment Setup} \label{sec:exp_setup}
\textbf{Datasets}. We benchmark on two datasets: P3M-10k \cite{li2021privacy} and PPM-100 \cite{ke2022modnet}. P3M-10k is by far the largest human portrait matting dataset and contains 10421 high-resolution in-the-wild images with annotated alpha mattes. For privacy issues, all faces in the images have been blurred. As shown in \cite{li2021privacy}, training on images with blurred faces does not degrade the model performance. Instead, it may even help the model to generalize better. We use the provided 9421 images with blurred faces for training and 500 images with blurred faces for privacy-preserving test and rest 500 normal images (without face blurring) for non-privacy test. Following \cite{li2021privacy}, we denote the two test subsets from P3M-10k as P3M-500-P (privacy-preserving) and P3M-500-NP (non-privacy). Compared to P3M-10k, PPM-100 is a smaller dataset curated specifically for evaluation purpose.

\textbf{Baselines}. We compare our model with the state-of-the-art trimap free method P3M-Net \cite{li2021privacy}. As a reference, we also include DIM \cite{xu2017deep}, a commonly adopted trimap-based baseline, in the experiments. We also compare with MODNet \cite{ke2022modnet} and BGMv2 \cite{lin2021real}, which are designated lightweight model for fast inference. The original BGMv2 relies on an additional image of background. To make it a fair comparison, we retrain a modified version by eliminating the background capture. Note that MODNet is designed for 512$\times$512 input images while we are targeting at higher resolutions such as HD and 4K. Therefore, we use two strategies to accommodate MODNet in our test scenario -- we either use 512$\times$512 input and upsample the output to full resolution or run the model on full resolution directly.      

\textbf{Default model}. Our default model uses Swin-T \cite{liu2021swin} as the low resolution network. To reduce the input image size for coarse alpha estimation, we apply a patch size of 16 for pixel-unshuffle, which is equivalent to a $\times$16 reduction in spatial resolution. Because the refinement network operates on 8$\times$8 regions, we let the low resolution network's decoder to produce a 4-channel output at $\mathfrak{R}_{16}$ and append a pixel-shuffle layer at the end to increase the resolution from $\mathfrak{R}_{16}$ to $\mathfrak{R}_8$. This way, a pixel at $\mathfrak{R}_8$ is equivalent to a 8$\times$8 region at the full resolution. For CRA, we use 8 nearest neighbors and employ a search range of 4 for relative positional encoding. 

\textbf{Evaluation metrics}. We follow previous works using the sum of absolute difference (SAD), the gradient difference (Grad) and the connectivity error (Conn) as the evaluation metrics. Conn is used as a way to measure the degree of connectivity, the intuition behind which is that unconnected components are more visually distracting when they are further away from the dominant connected components in the image \cite{rhemann2009perceptually}. We also report SAD within the transition area (a.k.a the ``uncertain'' region in a trimap), denoted as SAD-T. FLOPS is used as an indicator of compute budget.
Since the FLOPS of BGMv2 \cite{lin2021real} and our method depend
on image content, we report the mean FLOPS over multiple
inferences with an average of 1.63M pixels per inference.

\begin{figure}[h]
  \centering
  \includegraphics[width=0.48\textwidth]{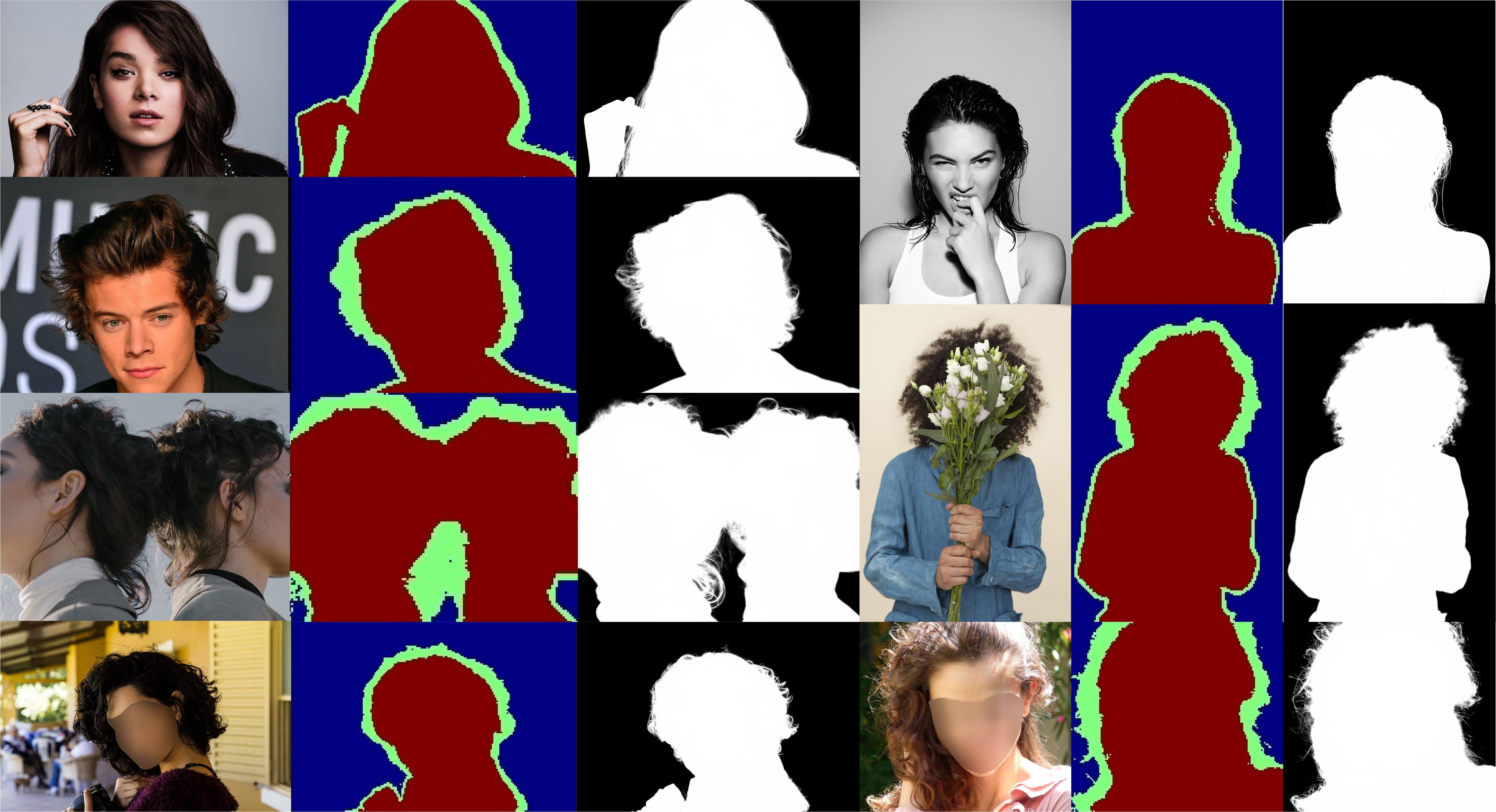}
  \caption{Estimated coarse trimaps and refined alpha mattes. Trimaps are upsampled to full resolution for visualization.}
  \label{fig:trimap_viz}
\end{figure}

\begin{figure*}[th]
  \centering
  \includegraphics[width=\textwidth]{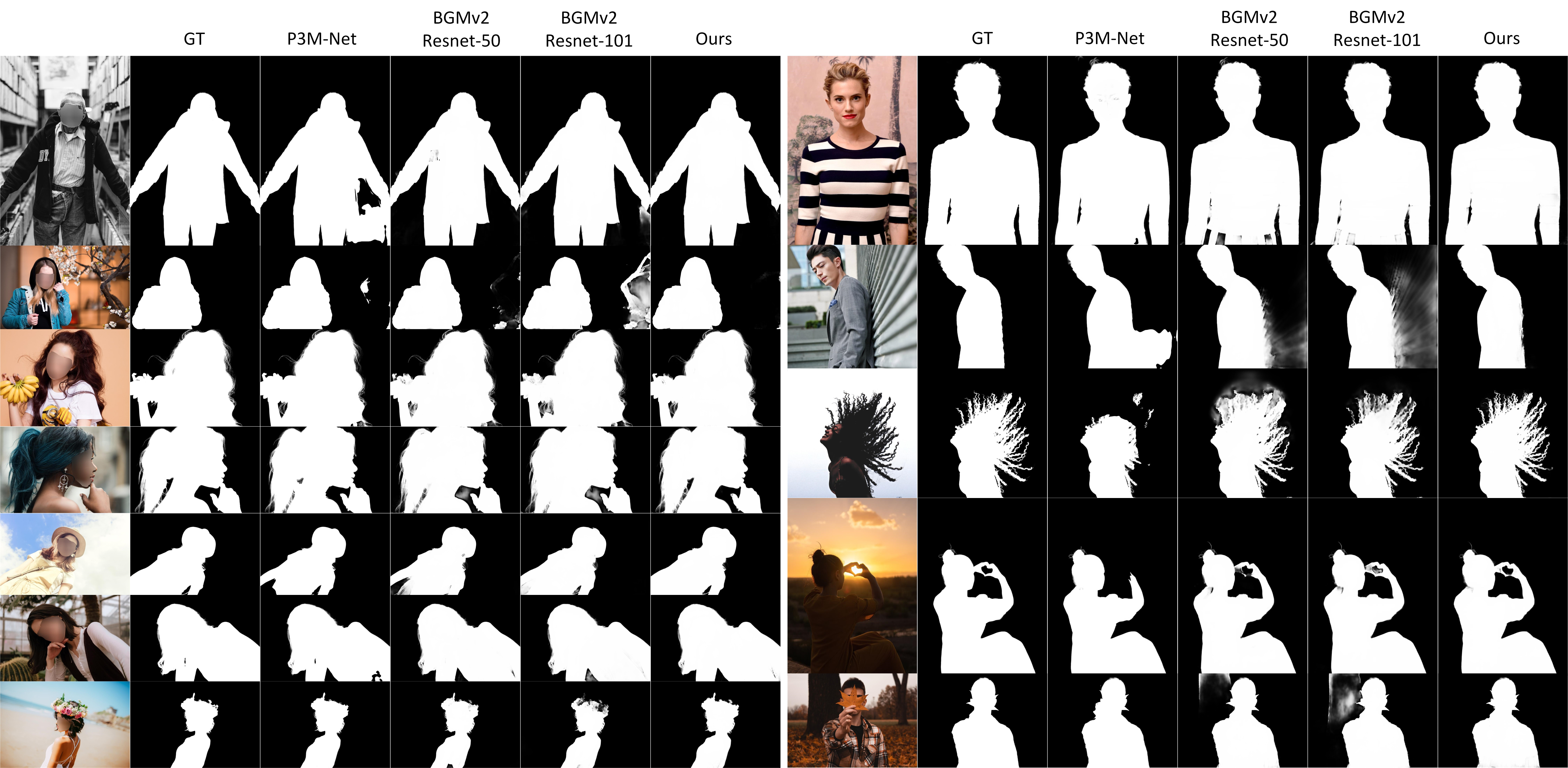}
  \caption{Qualitative results. On the left and right are respectively the results of P3M-500-P and P3M-500-NP. Zoom in for more details.}
  \label{fig:all_viz}
\end{figure*}
%\vspace{-0.02\hsize}

\subsection{Results}
We visualize the intermediate coarse trimaps and the final alpha mattes in Fig. \ref{fig:trimap_viz}. One can see how accurately our model can adapt to the number of regions to be refined. For example, the low resolution network predicts more ``uncertain'' regions (shown in green) around fuzzy hair in the trimaps while refrains from doing so around the contour of body. More qualitative results are shown in Fig. \ref{fig:all_viz}.

Quantitative results on the P3M test data are presented in Tab. \ref{tab:main_exp}. Our model outperforms all baselines by a large margin on all metrics while uses the least amount of computation. Compared to the previous state-of-the-art method P3M-Net, our model achieves competitive results with nearly $1/20$ of the FLOPS. We are only slightly behind P3M-Net on Grad (for both P3M-500-NP and P3M-500-P) and SAD (for P3M-500-P) while obtains the state-of-the-art results on all other metrics. In the auxiliary-free setting, BGMv2 does not retain the good performance reported by \cite{lin2021real} due to the lack of an additional background capture as input. Note that DIM has the best numbers for all metrics, but it is not directly comparable to other models because it takes trimap as an auxiliary input. We include it here merely for reference purpose as it is one of the mostly adopted methods for comparison.

It is worth noting that MODNet is originally designed and trained for 512x512 images. When running on high resolution input, not only does it incur degraded quality, but also increase its GFLOPS from 15.7 to 103.2. On the other hand, our model is super lightweight and can generate high quality full resolution mattes with only 19 FLOPS. 

Numeric results on the PPM-100 test data are shown in Tab. \ref{tab:PPM_exp}. Since PPM-100 does not have training data, we use models trained on the P3M-10k data for evaluation. Our model is superior to others on all metrics except being slightly worse than P3M-Net on SAD-T. P3M-Net achieve competitive results on P3M-500-NP and P3M-500-P, but its performance drops significantly when evaluated on PPM-100. We believe this is because of the domain gap between the training and test sets. Some major differences we observe between the two datasets are image resolutions and imaging quality. Images in PPM-100 are higher resolutions but have worse imaging quality. This explains the overall performance degradation for all models on PPM-100. However, our model is more robust to this domain gap and achieves the best results on PPM-100. 

\setlength{\tabcolsep}{0.008\textwidth} % set cell horizontal padding
\begin{table}[h]
    \small
    \centering
    \caption{Quantitative results on the PPM-100 dataset.}
    \label{tab:PPM_exp}
    \begin{tabular}{c|c|c|c|c|c}
    \toprule
    \multicolumn{2}{c|}{Method} & SAD & SAD-T & Grad & Conn \\
    \hline
    \multicolumn{2}{c|}{P3M-Net \cite{li2021privacy}} & 142.74 & 43.06 & 57.02 & 139.89 \\
    \hline
    \multirow{2}{*}{MODNet \cite{ke2022modnet}} & 512x512 & 104.35 & 65.42 & 68.56 & 96.45 \\
    \cline{2-6}
    & fullres & 324.07 & 68.97 & 77.42 & 319.70 \\
    \hline
    % e23/300k iters
    \multirow{2}{*}{BGMv2 \cite{lin2021real}} & Resnet-50 & 193.40 & 49.39 & 61.49 & 185.52 \\
    \cline{2-6}
    % e24_1/600k iters
    & Resnet-101 & 159.44 & 50.67 & 59.41 & 149.79 \\
    \hline
    % h7
    \multicolumn{2}{c|}{Ours} & 90.28 & 45.06 & 50.69 & 84.09 \\
    \hline
    \end{tabular}
\end{table}

\begin{comment}

\setlength{\tabcolsep}{0.008\textwidth} % set cell horizontal padding
\begin{table}[h]
    \small
    \centering
    \caption{Quantitative results on the PhotoMatte-85 dataset.}
    \label{tab:PhotoMatte_exp}
    \begin{tabular}{c|c|c|c|c|c}
    \toprule
    \multicolumn{2}{c|}{Method} & SAD & SAD-T & Grad & Conn \\
    \hline
    \multicolumn{2}{c|}{P3M-Net \cite{li2021privacy}} & 67.16 & 16.74 & 28.60 & 64.97 \\
    \hline
    \multirow{2}{*}{MODNet \cite{ke2022modnet}} & 512x512 & &  &  & \\
    \cline{2-6}
    & fullres & 248.86 & 23.91 & 38.97 & 244.65 \\
    \hline
    % e23/300k iters
    \multirow{2}{*}{BGMv2 \cite{lin2021real}} & Resnet-50 & 96.96 & 24.83 & 34.80 & 89.78 \\
    \cline{2-6}
    % e24_1/600k iters
    & Resnet-101 & 84.47 & 25.73 & 31.07 & 79.99 \\
    \hline
    % h7
    \multicolumn{2}{c|}{Ours} & 49.33 & 19.39 & 21.41 & 45.35 \\
    \bottomrule
    \end{tabular}
\end{table}

\end{comment}

\setlength{\tabcolsep}{0.008\textwidth} % set cell horizontal padding
\begin{table}[h]
    \small
    \centering
    \caption{FPS and GFLOPS for HD and 4K inputs. All models are evaluated with a single Nvidia Quadro RTX 6000 GPU. An empty entry means we fail to evaluate the model due to out-of-memory error.} 
    \label{tab:perf}
    \begin{tabular}{c|c|c|c|c|c}
    \toprule
    \multicolumn{2}{c|}{} & \multicolumn{2}{c|}{HD} & \multicolumn{2}{c}{4K} \\
    \hline
    \multicolumn{2}{c|}{Method} & FPS & GFLOPS & FPS & GFLOPS \\
    \hline
    % LF
    %\multicolumn{2}{c|}{LF \cite{zhang2019late}} & 1.2 & 9106.0 & - & - \\ %& 37.9M \\
    %\hline
    % HATT
    %\multicolumn{2}{c|}{HATT \cite{qiao2020attention}} & - & - & - & - \\ %& 105.4M \\
    %\hline
    % SHM
    %\multicolumn{2}{c|}{SHM \cite{chen2018semantic}} & 7.5 & 2461.9 & - & - \\ %& 74.7M \\
    %\hline
    % AIM
    %\multicolumn{2}{c|}{AIM \cite{li2021deep}} & 6.9 & 617.2 & - & - \\ %& 55.3M \\
    %\hline
    % DIM
    \multicolumn{2}{c|}{DIM \cite{xu2017deep}} & 5.0 & 1007.1 & - & - \\ %& 25.6M \\
    \hline
    % P3M-Net
    \multicolumn{2}{c|}{P3M-Net \cite{li2021privacy}} & 9.2 & 463.5 & - & - \\ %& 39.5M \\
    \hline
    \multicolumn{2}{c|}{MODNet \cite{ke2022modnet}} & 15.0 & 123.4 & - & - \\ %& 6.5M \\
    \hline
    % e23/300k iters
    BGMv2 & Resnet-50 & 57.4 & 32.7 & 23.7 & 128.6 \\ %& 40.2M \\
    \cline{2-6}
    % e24_1/600k iters
    \cite{lin2021real} & Resnet-101 & 45.8 & 42.2 & 17.8 & 166.8 \\ %& 59.2M \\
    \hline
    % h7
    \multirow{2}{*}{Ours} & w/ CRA & 54.9 & 21.2 & 19.5 & 74.6 \\ %& 31.3M \\
    \cline{2-6}
    % h17_4
    & w/o CRA & 71.2 & 19.4 & 26.4 & 70.7 \\ %& 31.0M \\
    \bottomrule
    \end{tabular}
\end{table}

\newcommand{\cmark}{\ding{51}}  % define check mark
\newcommand{\xmark}{\ding{55}}  % define cross mark
\setlength{\tabcolsep}{0.012\textwidth} % set cell horizontal
\begin{table*}[th]
    \small
    \centering
    \caption{Quantitative results for ablation study.}
    \label{tab:ablation}
    \begin{tabular}{c|c|c|c|c|c|c|c|c|c|c}
    \toprule
    \multicolumn{3}{c|}{}  & \multicolumn{4}{c|}{P3M-500-NP} & \multicolumn{4}{c}{P3M-500-P} \\
    \hline
    No. & Refinement Method & CRA & SAD & SAD-T & Grad & Conn & SAD & SAD-T & Grad & Conn \\
    \hline
    % h16_1/500k iters
    E & \cite{lin2021real} & NA &
    11.68 & 7.92 & 12.90 & 11.11 & 11.81 & 7.37 & 15.19 & 11.44 \\
    \hline
    % h17_4/600k iters
    F & Ours & \xmark &
    11.71 & 7.28 & 11.72 & 10.88 & 10.69 & 6.90 & 13.74 & 10.06 \\
    \hline
    % h7/500k iters
    G & Ours & \cmark & 10.60 & 6.83 & 10.78 & 9.77 & 10.04 & 6.44 & 12.65 & 9.41 \\
    \midrule
    \midrule
     & Search Range & KNN & \multicolumn{8}{c}{} \\
    \hline
    I & 2 & 8 & 11.58 & 6.82 & 11.04 & 10.76 & 10.52 & 6.48 & 13.06 & 9.87 \\
    \hline
    H & 3 & 8 & 10.74 & 6.85 & 10.78 & 9.91 & 10.41 & 6.46 & 12.76 & 9.77 \\
    \hline
    G & 4 & 8 & 10.60 & 6.83 & 10.78 & 9.77 & 10.04 & 6.44 & 12.65 & 9.41 \\
    \hline 
    J & 8 & 8 & 10.60 & 6.79 & 10.87 & 9.79 & 10.46 & 6.56 & 13.08 & 9.88 \\
    \hline
    K & 4 & 4 & 11.11 & 6.96 & 11.20 & 10.28 & 10.40 & 6.54 & 13.04 & 9.81 \\
    \hline 
    L & 4 & 16 & 10.77 & 6.80 & 10.93 & 9.95 & 9.40 & 6.33 & 12.59 & 8.84 \\
    \bottomrule
    \end{tabular}
\end{table*}

\begin{figure}[th]
  \centering
  \includegraphics[width=0.5\textwidth]{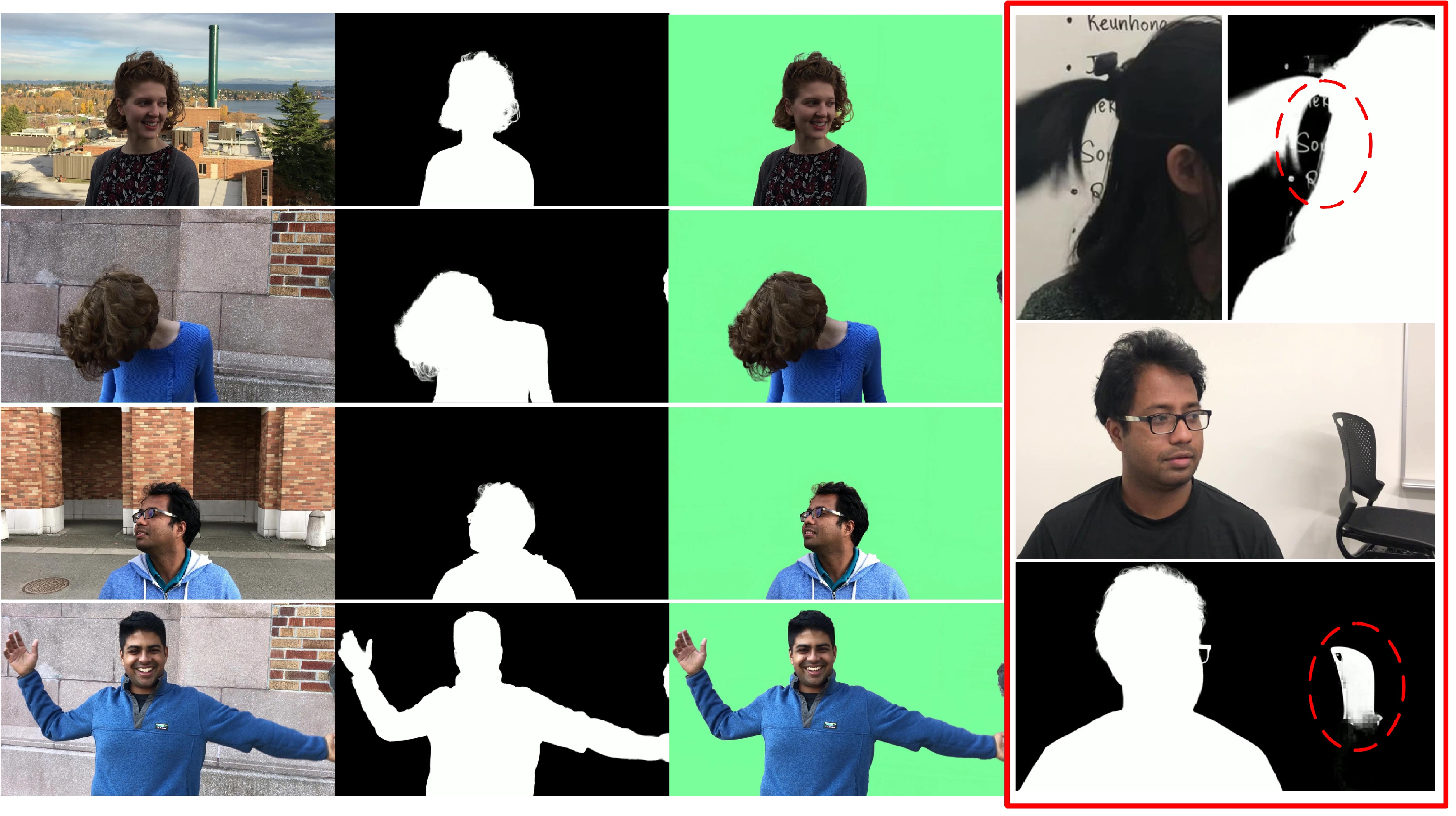}
  \caption{Qualitative results of real-world HD videos. The red box shows some of the typical failure cases.}
  \label{fig:real_video_w_fail}
\end{figure}

\begin{comment}
\begin{figure}[th]
  \centering
  \includegraphics[width=0.4\textwidth]{figs/real_video.jpg}
  \caption{XXX}
  \label{fig:real_video}
\end{figure}

\begin{figure}[th]
  \centering
  \includegraphics[width=0.4\textwidth]{figs/fail_case.jpg}
  \caption{XXX}
  \label{fig:fail_case}
\end{figure}
\end{comment}

\subsection{Real Application Performance}
We test on real-world HD videos from \cite{sengupta2020background,lin2021real} and show the qualitative results in Fig. \ref{fig:real_video_w_fail}. Please refer to the supplementary material for more video results. 

We profile all models on HD and 4K inputs and compare their FPS and GFLOPS in Tab. \ref{tab:perf}. As shown in the table, our models use the least amount of computation and achieve competitive frame rate. For DIM, P3M-Net and MODNet running on full resolution, we fail to profile their performance on 4K input due to the massive memory footprint required. On the other hand, our models yield real time performance for HD input and near real time for 4K. It is also worth noting that our model with CRA, even with fewer GFLOPS, runs slightly slower than Resnet-50 backboned BGMV2. This is because most modern deep learning frameworks do not have well optimized transformer operators. As shown in \cite{bettertransformer}, more than 2x speedup is possible with the optimized native CUDA kernels. Our current implementation of Swin-T and CRA utilizes generic pytorch functions to compute the multi-head attention. We believe a similar improvement at inference time is possible with the optimized implementation.

\subsection{Ablation Study}
In this section, we demonstrate the effectiveness of the proposed refinement stage, the CRA module and their associated hyper parameters.

\textbf{Refinement stage}. We compare the proposed refinement stage with that of BGMv2 \cite{lin2021real} by fixing the low resolution backbone. As shown in Tab. \ref{tab:ablation}, model G (our default) and model E differ only by the refinement stage. Model G outperforms model E on all metrics, demonstrating the effectiveness of the proposed refinement stage. We also observe similar results by comparing model A (from Tab. \ref{tab:trial_exp}) with BGMv2 Resnet-50 (from Tab. \ref{tab:main_exp}). Both models use Resnet-50 as the low resolution network, but they differ by the refinement stage. Model A surpassing BGMv2 Resnet-50 on all evaluated metrics, again, demonstrates the advantage of the proposed refinement stage.

\textbf{Cross region attention}. The CRA lies at the core of our method. We show its individual impact by taking it out from our model, resulting in model F shown in Tab. \ref{tab:ablation}. We can see that G achieves better results than F, demonstrating the effectiveness of CRA. 

\textbf{Search range}. Search range is used in the CRA module to identify in-range neighbors and out-of-range neighbors for relative positional encoding (Sec. \ref{sec:attn}). Because all out-of-range neighbors share the same relative positional bias (Fig. \ref{fig:pos_bias}), smaller search range enforces more out-of-range neighbors, making the model less discriminative against the neighboring regions. In Tab. \ref{tab:ablation}, we list three models (G, H \& I) with different search ranges. As the search range increases, we can see a trend \footnote[1]{For the majority of the evaluated metrics.} in improved performance. 

However, we also observe from model J that a large search range of 8 does not boost the performance any further. We believe this is due to the search window being too large (15$\times$15) \footnote[2]{Recall that, given the search range $s$, the search window size is $(2s-1)\times(2s-1)$.}. At most 8 out of 225 positional biases are queried from the lookup table, leaving the rest of the positional biases untouched. Therefore, a small percentage of biases being queried and optimized during each gradient propagation results in sub-optimal bias lookup table, hence the degraded quality of the model.

\textbf{KNNs}. We study the impact of KNNs by varying its number. As KNNs increase (K $\rightarrow$ G $\rightarrow$ L) in Tab. \ref{tab:ablation}, we can see the overall \footnotemark[1] performance of the model improves. The reason is twofold. First, more KNNs means more contextual information, which helps the model do a better job at learning. Second, more KNNs benefit the training of bias lookup table. As we have just discussed, it improves the chances of positional biases being queried and optimized during training.

\section{Failure cases and Future Work}
When there is high contrast texture in the extracted regions, the refinement network finds it difficult to identify the correct foreground and background. As is shown in Fig. \ref{fig:real_video_w_fail}, the text in on the whiteboard is supposed to be background, but it is perfectly (and incorrectly) segmented as foreground. Also, the refinement network can only improvement the quality of local boundaries. Any false predictions in the original low resolution matte can not be recovered. For example, The chair in Fig. \ref{fig:real_video_w_fail} has been false positive in the first stage, it is impossible to undo the false prediction by the refinement network. 

Currently our model is trained only with limited amount of data (9421 images from the P3M data \cite{li2021privacy}), which is far from being robust in real-world applications. Because the refinement network only consumes an upsampled matte and does not rely on any intermediate features from the first stage, we believe it is possible to train the low resolution network and the refinement network separately to improvement the overall robustness of our method. The abundant low resolution segmentation data \cite{lin2014microsoft, zhao2018understanding, li2017towards} can be leveraged to train the coarse model while the high resolution matting data plus an unlimited number of human synthetics \cite{wood2021fake} can used to train the refinement stage. We leave this as a future work.

\section{Conclusion}
We present a new lightweight two-stage method for high resolution portrait matting. At the heart of our method is a ViT backboned low resolution network for coarse alpha estimation and a novel cross region attention (CRA) module in the second stage for local refinement. We verify that using pixel-unshuffle rather than downsampling has the advantage of preserving original pixel information and that ViT comes naturally a good choice for that purpose. We demonstrate the effectiveness of the proposed low resolution network, refinement stage and CRA module and analyze the individual impact of several key hyper parameters. Through extensive experiments, we show the superiority of our method against the previous state-of-the-arts in terms of both accuracy, FPS and FLOPS.

%%%%%%%%% REFERENCES
{\small
\bibliographystyle{ieee_fullname}
\bibliography{egbib}

\begin{thebibliography}{10}\itemsep=-1pt

\bibitem{aksoy2017designing}
Yagiz Aksoy, Tunc Ozan~Aydin, and Marc Pollefeys.
\newblock Designing effective inter-pixel information flow for natural image
  matting.
\newblock In {\em Proceedings of the IEEE Conference on Computer Vision and
  Pattern Recognition}, pages 29--37, 2017.

\bibitem{chen2018semantic}
Quan Chen, Tiezheng Ge, Yanyu Xu, Zhiqiang Zhang, Xinxin Yang, and Kun Gai.
\newblock Semantic human matting.
\newblock In {\em Proceedings of the 26th ACM international conference on
  Multimedia}, pages 618--626, 2018.

\bibitem{chen2013knn}
Qifeng Chen, Dingzeyu Li, and Chi-Keung Tang.
\newblock Knn matting.
\newblock {\em IEEE transactions on pattern analysis and machine intelligence},
  35(9):2175--2188, 2013.

\bibitem{chuang2001bayesian}
Yung-Yu Chuang, Brian Curless, David~H Salesin, and Richard Szeliski.
\newblock A bayesian approach to digital matting.
\newblock In {\em Proceedings of the 2001 IEEE Computer Society Conference on
  Computer Vision and Pattern Recognition. CVPR 2001}, volume~2, pages II--II.
  IEEE, 2001.

\bibitem{dosovitskiy2020image}
Alexey Dosovitskiy, Lucas Beyer, Alexander Kolesnikov, Dirk Weissenborn,
  Xiaohua Zhai, Thomas Unterthiner, Mostafa Dehghani, Matthias Minderer, Georg
  Heigold, Sylvain Gelly, et~al.
\newblock An image is worth 16x16 words: Transformers for image recognition at
  scale.
\newblock {\em arXiv preprint arXiv:2010.11929}, 2020.

\bibitem{forte2020f}
Marco Forte and Fran{\c{c}}ois Piti{\'e}.
\newblock $ f $, $ b $, alpha matting.
\newblock {\em arXiv preprint arXiv:2003.07711}, 2020.

\bibitem{he2010fast}
Kaiming He, Jian Sun, and Xiaoou Tang.
\newblock Fast matting using large kernel matting laplacian matrices.
\newblock In {\em 2010 IEEE Computer Society Conference on Computer Vision and
  Pattern Recognition}, pages 2165--2172. IEEE, 2010.

\bibitem{he2016deep}
Kaiming He, Xiangyu Zhang, Shaoqing Ren, and Jian Sun.
\newblock Deep residual learning for image recognition.
\newblock In {\em Proceedings of the IEEE conference on computer vision and
  pattern recognition}, pages 770--778, 2016.

\bibitem{he2016identity}
Kaiming He, Xiangyu Zhang, Shaoqing Ren, and Jian Sun.
\newblock Identity mappings in deep residual networks.
\newblock In {\em European conference on computer vision}, pages 630--645.
  Springer, 2016.

\bibitem{ke2022modnet}
Zhanghan Ke, Jiayu Sun, Kaican Li, Qiong Yan, and Rynson~WH Lau.
\newblock Modnet: Real-time trimap-free portrait matting via objective
  decomposition.
\newblock In {\em Proceedings of the AAAI Conference on Artificial
  Intelligence}, volume~36, pages 1140--1147, 2022.

\bibitem{levin2007closed}
Anat Levin, Dani Lischinski, and Yair Weiss.
\newblock A closed-form solution to natural image matting.
\newblock {\em IEEE transactions on pattern analysis and machine intelligence},
  30(2):228--242, 2007.

\bibitem{li2021privacy}
Jizhizi Li, Sihan Ma, Jing Zhang, and Dacheng Tao.
\newblock Privacy-preserving portrait matting.
\newblock In {\em Proceedings of the 29th ACM International Conference on
  Multimedia}, pages 3501--3509, 2021.

\bibitem{li2017towards}
Jianshu Li, Jian Zhao, Yunchao Wei, Congyan Lang, Yidong Li, Terence Sim,
  Shuicheng Yan, and Jiashi Feng.
\newblock Multi-human parsing in the wild.
\newblock {\em arXiv preprint arXiv:1705.07206}, 2017.

\bibitem{li2020natural}
Yaoyi Li and Hongtao Lu.
\newblock Natural image matting via guided contextual attention.
\newblock In {\em Proceedings of the AAAI Conference on Artificial
  Intelligence}, volume~34, pages 11450--11457, 2020.

\bibitem{lin2021real}
Shanchuan Lin, Andrey Ryabtsev, Soumyadip Sengupta, Brian~L Curless, Steven~M
  Seitz, and Ira Kemelmacher-Shlizerman.
\newblock Real-time high-resolution background matting.
\newblock In {\em Proceedings of the IEEE/CVF Conference on Computer Vision and
  Pattern Recognition}, pages 8762--8771, 2021.

\bibitem{lin2014microsoft}
Tsung-Yi Lin, Michael Maire, Serge Belongie, James Hays, Pietro Perona, Deva
  Ramanan, Piotr Doll{\'a}r, and C~Lawrence Zitnick.
\newblock Microsoft coco: Common objects in context.
\newblock In {\em Computer Vision--ECCV 2014: 13th European Conference, Zurich,
  Switzerland, September 6-12, 2014, Proceedings, Part V 13}, pages 740--755.
  Springer, 2014.

\bibitem{liu2021swin}
Ze Liu, Yutong Lin, Yue Cao, Han Hu, Yixuan Wei, Zheng Zhang, Stephen Lin, and
  Baining Guo.
\newblock Swin transformer: Hierarchical vision transformer using shifted
  windows.
\newblock In {\em Proceedings of the IEEE/CVF International Conference on
  Computer Vision}, pages 10012--10022, 2021.

\bibitem{lu2019indices}
Hao Lu, Yutong Dai, Chunhua Shen, and Songcen Xu.
\newblock Indices matter: Learning to index for deep image matting.
\newblock In {\em Proceedings of the IEEE/CVF International Conference on
  Computer Vision}, pages 3266--3275, 2019.

\bibitem{lutz2018alphagan}
Sebastian Lutz, Konstantinos Amplianitis, and Aljosa Smolic.
\newblock Alphagan: Generative adversarial networks for natural image matting.
\newblock {\em arXiv preprint arXiv:1807.10088}, 2018.

\bibitem{bettertransformer}
Scott Wolchok Rui Zhu Christian~Puhrsch Michael~Gschwind, Eric~Han.
\newblock A bettertransformer for fast transformer inference.
\newblock
  \url{https://pytorch.org/blog/a-better-transformer-for-fast-transformer-encoder-inference/}.

\bibitem{qiao2020attention}
Yu Qiao, Yuhao Liu, Xin Yang, Dongsheng Zhou, Mingliang Xu, Qiang Zhang, and
  Xiaopeng Wei.
\newblock Attention-guided hierarchical structure aggregation for image
  matting.
\newblock In {\em Proceedings of the IEEE/CVF Conference on Computer Vision and
  Pattern Recognition}, pages 13676--13685, 2020.

\bibitem{rhemann2009perceptually}
Christoph Rhemann, Carsten Rother, Jue Wang, Margrit Gelautz, Pushmeet Kohli,
  and Pamela Rott.
\newblock A perceptually motivated online benchmark for image matting.
\newblock In {\em 2009 IEEE conference on computer vision and pattern
  recognition}, pages 1826--1833. IEEE, 2009.

\bibitem{sengupta2020background}
Soumyadip Sengupta, Vivek Jayaram, Brian Curless, Steven~M Seitz, and Ira
  Kemelmacher-Shlizerman.
\newblock Background matting: The world is your green screen.
\newblock In {\em Proceedings of the IEEE/CVF Conference on Computer Vision and
  Pattern Recognition}, pages 2291--2300, 2020.

\bibitem{simonyan2014very}
Karen Simonyan and Andrew Zisserman.
\newblock Very deep convolutional networks for large-scale image recognition.
\newblock {\em arXiv preprint arXiv:1409.1556}, 2014.

\bibitem{sun2004poisson}
Jian Sun, Jiaya Jia, Chi-Keung Tang, and Heung-Yeung Shum.
\newblock Poisson matting.
\newblock In {\em ACM SIGGRAPH 2004 Papers}, pages 315--321. 2004.

\bibitem{szegedy2015going}
Christian Szegedy, Wei Liu, Yangqing Jia, Pierre Sermanet, Scott Reed, Dragomir
  Anguelov, Dumitru Erhan, Vincent Vanhoucke, and Andrew Rabinovich.
\newblock Going deeper with convolutions.
\newblock In {\em Proceedings of the IEEE conference on computer vision and
  pattern recognition}, pages 1--9, 2015.

\bibitem{vaswani2017attention}
Ashish Vaswani, Noam Shazeer, Niki Parmar, Jakob Uszkoreit, Llion Jones,
  Aidan~N Gomez, {\L}ukasz Kaiser, and Illia Polosukhin.
\newblock Attention is all you need.
\newblock {\em Advances in neural information processing systems}, 30, 2017.

\bibitem{wood2021fake}
Erroll Wood, Tadas Baltru{\v{s}}aitis, Charlie Hewitt, Sebastian Dziadzio,
  Thomas~J Cashman, and Jamie Shotton.
\newblock Fake it till you make it: face analysis in the wild using synthetic
  data alone.
\newblock In {\em Proceedings of the IEEE/CVF international conference on
  computer vision}, pages 3681--3691, 2021.

\bibitem{xu2017deep}
Ning Xu, Brian Price, Scott Cohen, and Thomas Huang.
\newblock Deep image matting.
\newblock In {\em Proceedings of the IEEE conference on computer vision and
  pattern recognition}, pages 2970--2979, 2017.

\bibitem{zhang2019late}
Yunke Zhang, Lixue Gong, Lubin Fan, Peiran Ren, Qixing Huang, Hujun Bao, and
  Weiwei Xu.
\newblock A late fusion cnn for digital matting.
\newblock In {\em Proceedings of the IEEE/CVF conference on computer vision and
  pattern recognition}, pages 7469--7478, 2019.

\bibitem{zhang2018residual}
Yulun Zhang, Yapeng Tian, Yu Kong, Bineng Zhong, and Yun Fu.
\newblock Residual dense network for image super-resolution.
\newblock In {\em Proceedings of the IEEE conference on computer vision and
  pattern recognition}, pages 2472--2481, 2018.

\bibitem{zhao2018understanding}
Jian Zhao, Jianshu Li, Yu Cheng, Li Zhou, Terence Sim, Shuicheng Yan, and
  Jiashi Feng.
\newblock Understanding humans in crowded scenes: Deep nested adversarial
  learning and a new benchmark for multi-human parsing.
\newblock {\em arXiv preprint arXiv:1804.03287}, 2018.

\end{thebibliography}


\begin{thebibliography}{10}\itemsep=-1pt

\bibitem{chen2018semantic}
Quan Chen, Tiezheng Ge, Yanyu Xu, Zhiqiang Zhang, Xinxin Yang, and Kun Gai.
\newblock Semantic human matting.
\newblock In {\em Proceedings of the 26th ACM international conference on
  Multimedia}, pages 618--626, 2018.

\bibitem{ke2022modnet}
Zhanghan Ke, Jiayu Sun, Kaican Li, Qiong Yan, and Rynson~WH Lau.
\newblock Modnet: Real-time trimap-free portrait matting via objective
  decomposition.
\newblock In {\em Proceedings of the AAAI Conference on Artificial
  Intelligence}, volume~36, pages 1140--1147, 2022.

\bibitem{levin2007closed}
Anat Levin, Dani Lischinski, and Yair Weiss.
\newblock A closed-form solution to natural image matting.
\newblock {\em IEEE transactions on pattern analysis and machine intelligence},
  30(2):228--242, 2007.

\bibitem{li2021privacy}
Jizhizi Li, Sihan Ma, Jing Zhang, and Dacheng Tao.
\newblock Privacy-preserving portrait matting.
\newblock In {\em Proceedings of the 29th ACM International Conference on
  Multimedia}, pages 3501--3509, 2021.

\bibitem{li2021deep}
Jizhizi Li, Jing Zhang, and Dacheng Tao.
\newblock Deep automatic natural image matting.
\newblock {\em arXiv preprint arXiv:2107.07235}, 2021.

\bibitem{lin2021real}
Shanchuan Lin, Andrey Ryabtsev, Soumyadip Sengupta, Brian~L Curless, Steven~M
  Seitz, and Ira Kemelmacher-Shlizerman.
\newblock Real-time high-resolution background matting.
\newblock In {\em Proceedings of the IEEE/CVF Conference on Computer Vision and
  Pattern Recognition}, pages 8762--8771, 2021.

\bibitem{lin2017focal}
Tsung-Yi Lin, Priya Goyal, Ross Girshick, Kaiming He, and Piotr Doll{\'a}r.
\newblock Focal loss for dense object detection.
\newblock In {\em Proceedings of the IEEE international conference on computer
  vision}, pages 2980--2988, 2017.

\bibitem{qiao2020attention}
Yu Qiao, Yuhao Liu, Xin Yang, Dongsheng Zhou, Mingliang Xu, Qiang Zhang, and
  Xiaopeng Wei.
\newblock Attention-guided hierarchical structure aggregation for image
  matting.
\newblock In {\em Proceedings of the IEEE/CVF Conference on Computer Vision and
  Pattern Recognition}, pages 13676--13685, 2020.

\bibitem{xu2017deep}
Ning Xu, Brian Price, Scott Cohen, and Thomas Huang.
\newblock Deep image matting.
\newblock In {\em Proceedings of the IEEE conference on computer vision and
  pattern recognition}, pages 2970--2979, 2017.

\bibitem{zhang2019late}
Yunke Zhang, Lixue Gong, Lubin Fan, Peiran Ren, Qixing Huang, Hujun Bao, and
  Weiwei Xu.
\newblock A late fusion cnn for digital matting.
\newblock In {\em Proceedings of the IEEE/CVF conference on computer vision and
  pattern recognition}, pages 7469--7478, 2019.

\end{thebibliography}
}

\end{document}

\title{Lightweight Portrait Matting via Regional Attention and Refinement
Supplementary Material
}

\maketitle

%\FloatBarrier
\setlength{\tabcolsep}{0.015\textwidth} % set cell horizontal
\begin{table*}[b!]
    \small
    \centering
    \caption{Quantitative results on the P3M-500 tet data with additional baselines, which are listed in the upper part of the table. The lower part is copied from Table 2 of the main paper. $\dagger$ indicates that a trimap is used.}
    \label{tab:supp_exp}
    \begin{tabular}{c|c|c|c|c|c|c|c|c|c|c}
    \toprule
    \multicolumn{3}{c|}{}  & \multicolumn{4}{c|}{P3M-500-NP} & \multicolumn{4}{c}{P3M-500-P} \\
    \hline
    \multicolumn{2}{c|}{Method} & GFLOPS & SAD & SAD-T & Grad & Conn & SAD & SAD-T & Grad & Conn \\
    \hline
    \multicolumn{2}{c|}{LF \cite{zhang2019late}} & 7190.0 & 32.59 & 14.53 & 31.93 & 19.50 & 42.95 & 12.43 & 42.19 & 18.80  \\
    \hline
    \multicolumn{2}{c|}{HATT \cite{qiao2020attention}} & 4264.3 & 30.53 & 13.48 & 19.88 & 27.42 & 25.99 & 11.03 & 14.91 & 25.29 \\
    \hline
    \multicolumn{2}{c|}{SHM \cite{chen2018semantic}} & 1943.3 & 20.77 & 9.14 & 20.30 & 17.09 & 21.56 & 9.14 & 21.24 & 17.53 \\
    \hline
    \multicolumn{2}{c|}{AIM \cite{li2021deep}} & 487.4 & 15.50 & 10.16 & 14.82 & 18.03 & 13.20 & 8.84 & 12.58 & 17.75 \\
    \midrule
    \multicolumn{11}{c}{} \\
    \midrule
    \multicolumn{2}{c|}{DIM$^\dagger$ \cite{xu2017deep}} & 791.6 & 5.32 & 5.32 & 4.70 & 7.70 & 4.89 & 4.89 & 4.48 & 9.68 \\
    \hline
    \multicolumn{2}{c|}{P3M-Net \cite{li2021privacy}} & 364.9 & 11.23 & 7.65 & 10.35 & 12.51 & 8.73 & 6.89 & 8.22 & 13.88 \\
    \hline
    \multirow{2}{*}{MODNet \cite{ke2022modnet}} & 512x512 input & 15.7 & 20.20 & 12.48 & 16.83 & 18.41 & 30.08 & 12.22 & 19.73 & 28.61 \\
    \cline{2-11}
    & fullres input & 103.2 & 63.74 & 13.56 & 25.75 & 62.69 & 95.47 & 13.70 & 37.28 & 94.86 \\
    \hline
    % e22
    %\multirow{3}{*}{BGMv2 \cite{lin2021real}} & MobileNet-V2 & 7.7 & 21.27 & 8.16 & 14.16 & 18.91 & 21.28 & 7.58 & 17.06 & 19.48 \\
    %\cline{2-11}
    % e23/300k iters
    \multirow{2}{*}{BGMv2 \cite{lin2021real}} & Resnet-50 & 26.5 & 16.72 & 7.55 & 13.00 & 15.39 & 15.70 & 7.23 & 15.54 & 14.71 \\
    \cline{2-11}
    % e24_1/600k iters
    & Resnet-101 & 33.9 & 15.66 & 7.72 & 12.42 & 14.65 & 13.90 & 7.23 & 14.69 & 13.13 \\
    \hline
    % h7
    \multicolumn{2}{c|}{Ours} & 19.0 & 10.60 & 6.83 & 10.78 & 9.77 & 10.04 & 6.44 & 12.65 & 9.41 \\
    \bottomrule
    \end{tabular}
\end{table*}
%\FloatBarrier

\section{Training Losses}
The low resolution network is trained with binary cross entropy for the coarse alpha $\hat{\alpha}_c$ and focal loss \cite{lin2017focal} for the trimap $\hat{\tau}_c$ (one-hot encoded):
\begin{equation}
    \mathcal{L}_{\hat{\alpha}_c} = -\alpha_c \cdot log(\hat{\alpha}_c) - (1 - \alpha_c) \cdot log(1 - \hat{\alpha}_c),
\end{equation}
\begin{equation}
    \mathcal{L}_{\hat{\tau}_c} = -\sum_{i=1}^{3} w_i \cdot (1 - p_i)^2 \cdot \tau_{c,i} \cdot log(\hat{\tau}_{c,i}),
\end{equation}
\begin{equation}
    p_i=\tau_{c,i} \cdot \hat{\tau}_{c,i} + (1-\tau_{c,i}) \cdot (1-\hat{\tau}_{c,i}),
\end{equation}
where $\alpha_c$ and $\tau_c$ are the downsampled ground truth alpha and trimap; $i$ is the trimap class index; $p_i$ measures how close the prediction $\hat{\tau}_c$ is to the ground truth $\tau_c$ and $(1 - p_i)^2$ in the focal loss is designed to weigh down the well predicted samples; $w_i$ is a weight to handle class imbalance. Empirically we set $w_i=\frac{n_i^{-0.5}}{\sum_i n_i^{-0.5}}$ for class $i$ with $n_i$ pixels.  

In addition to computing $\mathcal{L}_{\hat{\alpha}_c}$ and $\mathcal{L}_{\hat{\tau}_c}$ at $\mathfrak{R}_8$ (the output low resolution), we also add two output heads at $\mathfrak{R}_{16}$ and $\mathfrak{R}_{32}$ and compute the losses accordingly. The two added output heads are discarded during inference.

The full resolution alpha $\hat{\alpha}$ from the refinement network is trained with alpha loss and Laplacian loss \cite{li2021privacy,lin2021real}:
\begin{equation}
    \mathcal{L}_{\hat{\alpha}} = \sqrt{(\alpha - \hat{\alpha})^2 + \epsilon^2},
\end{equation}
\begin{equation}
    \mathcal{L}_{lap} = \sum_{r} \sqrt{(\mathcal{P}_r(\alpha) - \mathcal{P}_r(\hat{\alpha}))^2 + \epsilon^2},
\end{equation}
where $\mathcal{P}_r(\cdot)$ creates a Laplacian pyramid at resolution $r$. We use a 5-level pyramid with resolutions from $\mathfrak{R}_1$ to $\mathfrak{R}_{32}$.

Follwoing \cite{xu2017deep,li2021privacy,lin2021real}, We also adopt a composition loss by computing the difference between the images composed using the ground truth $\alpha$ and the predicted $\hat{\alpha}$:
\begin{equation}
    \mathcal{L}_{comp} = \sqrt{(\mathcal{C}(\alpha) - \mathcal{C}(\hat{\alpha}))^2 + \epsilon^2},
\end{equation}
where $\mathcal{C}(\alpha)=\alpha F + (1 - \alpha) B$ composes a new image with the source foreground $F$, a new background $B$ and the alpha matte $\alpha$. The foreground $F$ is estimated offline using \cite{levin2007closed} given the generated pseudo ground truth trimap. The final loss is the sum of all the losses from above: $\mathcal{L} = \mathcal{L}_{\hat{\alpha}_c} + \mathcal{L}_{\hat{\tau}_c} +\mathcal{L}_{\hat{\alpha}} + \mathcal{L}_{lap} + \mathcal{L}_{comp}$.

\section{Additional Baselines}
Due to space limit, we show only the most representative baselines in the main paper. There are other popular baseline methods that we do not include in the main paper but are often used by prior works. We summarize the results in Tab. \ref{tab:supp_exp}. Tab. \ref{tab:supp_exp} extends Tab. 2 (in the main paper) with additional baselines such as LF \cite{zhang2019late}, HATT \cite{qiao2020attention}, SHM \cite{chen2018semantic}, and AIM \cite{li2021deep}. 
As one as see from the table, our model outperforms all the methods in the table. The added baselines do not change our conclusion since have included the best performing baselines (e.g. DIM and P3M-Net) in the main paper. We list them here for completeness and a more comprehensive comparison.

\section{Results on Real Videos}
Please check out the attached demo videos in the supplementary material.

% need to add a figure here to activate pdftex; otherwise it's using latex2e, which can't pass arxiv
\begin{figure}[t]
    \centering
    \includegraphics[width=\linewidth]{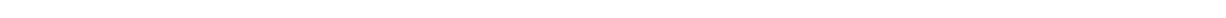}
    \label{fig:blank_fig}
\end{figure}

\clearpage

%%%%%%%%% REFERENCES
{\small
\bibliographystyle{ieee_fullname}
\bibliography{egbib}
}